# Image enhancement in intensity projected multichannel MRI using spatially adaptive directional anisotropic diffusion

**Akshara P Krishnan, Paul J S** *
Medical Image Computing and Signal Processing Laboratory, Indian Institute of Information Technology and Management- Kerala, Trivandrum – 695581, India
* +91-9846627817
* j.paul@iiitmk.ac.in

## Abstract

Anisotropic Diffusion is widely used for noise reduction with simultaneous preservation of vascular structures in maximum intensity projected (MIP) angiograms. However, extension to minimum intensity projected (mIP) venograms in Susceptibility Weighted Imaging (SWI) poses difficulties due to spatially varying baseline. Here, we introduce a modified version of the directional anisotropic diffusion which allows us to simultaneously reduce the noise and enhance vascular structures reconstructed using both M/mIP angiograms. This method is based on spatial adaptation of the diffusion function, separately in the directions of the gradient, and along those of the minimum and maximum curvatures. The existing approach of directional anisotropic diffusion uses binary switched diffusion function to ensure diffusion along the direction of maximum curvature stopped near the vessel borders. Here, the choice of a threshold for detecting the upper limit of diffusion becomes difficult in the presence of spatially varying baseline. Also, the approach of using vesselness measure to steer the diffusion process results in structural discontinuities due to junction suppression in mIP. The merits of the proposed method include elimination of the need for an *apriori* choice of a threshold to detect the vessel, and problems due to junction suppression. The proposed method is also extended to multi-channel phase contrast angiogram.

## 1. Introduction

In MR and CT images, vascular geometry is visualized using Intensity Projections. In Magnetic Resonance imaging, projection methods are used in both Time of Flight and Phase contrast (PC) angiography. The main artifacts associated with projection methods are background intensity variations and inherent noise in the signal prior to projection [1]. While performing projection, the noise component also gets projected. Since composition of intensity projected images have a mean base line level together with positive and negative swings, additive noise can easily obscure the swings closer to the base line. Since features of interest in intensity projected images consist of amplitude and width of the positive swings for Maximum Intensity Projection (MIP), and negative swings for Minimum Intensity Projection (mIP), it is required to retain and enhance the swing amplitudes during any pre-processing operation. Consequently, any denoising filter applied to the individual slices prior to projection should retain, and preferably enhance the swing amplitudes along with simultaneous noise reduction in the base line regions. Secondly, the denoising should be robust; i.e., noise removal method should also retain/enhance the swings closest to the noise level.

The taxonomy of denoising methods in medical imaging applications has evolved from the idea of nonlinear edge preserved smoothing filters. These filters are derived using either optimization based or diffusion based approaches. Of these, the former uses either distance based or energy based functionals. In the most elementary form, each of the above filtering method can be grouped into those based on Non Local Means (NLM) distances [2], Total Variation (TV) functional [3], and Anisotropic diffusion based filtering schemes [4]. Depending on the specific nature of the images and requirements of the denoising application, a cohort of methods using a combination of the elemental filtering forms has been evolved [5,6,7,8].

In principle, any edge preserving filter with denoising capability can be applied for localized noise reduction in intensity projected images. The most popular one is the Non-Local Means (NLM) filter first introduced by Buades et al. which exploits the natural redundancy of an image for structure preserved filtering. NLM method is highly dependent on setting of its parameters. There is a weight decay control parameter $h$ which controls the degree of filtering. If $h$ value is set too low, it will not denoise the image. In contrast, large $h$ values introduce additional blur. Hence, a direct application of NLM filter fails to maintain the swing amplitudes in M/mIP images. Likewise, TV method originally proposed by Rudin, Osher and Fatemi is a denoising filter which preserves the sharp edges. It is defined as an optimization problem based on minimizing a particular cost function. Any algorithm that solves the optimization problem can be used to implement TV denoising. Effectiveness of the TV method depends on choosing the value of smoothing parameter $\lambda$ [9]. It must be set low to remove noise, but at the cost of additional blur and loss of signal. So, there must be an effective means to choose the value of $\lambda$ when applied to pre-processing of intensity projected images. Also, TV method can sometimes introduce undesirable staircasing effect [10].

Anisotropic diffusion filtering, proposed by Perona and Malik [3], is a promising technique because of its simultaneous noise reduction and edge preserving properties. The smoothing process in this filter is simulated as a diffusive process that is controlled by weighting the diffusion rate with an exponential function of the signal intensity gradient. Application of this technique to intensity projected images can result in amplitude reduction of vascular intensities.

In the recent approach of Jiang Du et al. [11] for noise reduction using nonlinear anisotropic filtering, a composite image data set that combines the filtered low spatial frequencies and unaltered high spatial frequencies is first formed. This is then followed by application of nonlinear anisotropic filter. In this approach, smoothing parameter is determined from prior estimation of background noise. Apart from the difficulty associated with estimation of background noise, spatial frequency dependent nonlinear anisotropic filtering leads to reduction of amplitude swings in both positive and negative directions. Also, the number of iterations is subject to the local image characteristics and choice of smoothing parameter.

In the orthogonal decomposition of Perona malik diffusion [12], anisotropic diffusion can be viewed as two 1-D evolutions moving perpendicular to each other. In this, each component achieves smoothing along and perpendicular to the vascular edges. Furthermore, several works were proposed in which vessel architecture and orientation is determined by analysis of Hessian matrices [13,14,15,16]. Frangi et al. [14] introduced the vesselness measure based on eigen values extracted from the Hessian matrix in a multiscale fashion and one drawback of this method is that they include adjacent features e.g. heart chambers, or other organs as vessels. Krissian et al. [15] studied the relation between the Hessian matrix and the image gradient computed in multiple scales for the detection of tubular structures. Manniesing et al. [16] made use of the multiscale Hessian matrix based features to steer the diffusion using the vessel resemblance function.

In this paper, we discuss a diffusion based filtering, leading to increased swing amplitudes with concomitant denoising, applied to both flow compensated Susceptibility Weighted Imaging (SWI) and multi-channel phase contrast angiograms. The directional smoothing approach [15] is utilized for denoising of M/mIP images. We use a 3-D basis $\eta$, $e_1$ and $e_2$ similar to the one proposed in [15], which corresponds, respectively, to unit vectors in the directions of the gradient and the maximal and minimal curvatures. Smoothing is performed along these individual directions for better noise removal. Spatial adaptation of these directional derivatives increases the contrast of vascular edges. Since edges are well defined in MIP images, directional smoothing along the minimum curvature and gradient is sufficient.

The paper is organized as follows. The underlying principle of directional anisotropic diffusion filtering and steps leading to the proposed filtering scheme are provided in Section 2. The motivation and need for a spatially adaptive directional diffusion with applications to venous enhancement in SWI and phased array reconstruction of maximum intensity projected angiograms are also included in this section. Section 3 summarizes the experimental results along with a brief summary and discussion presented in Section 4.

# 2. Material and Methods

## 2.1 Background of Diffusion based Filter

### Anisotropic Diffusion

Anisotropic diffusion filtering (ADF) was first proposed by Perona and Malik [4] as nonlinear diffusion method which aims at reducing the image noise while keeping the image information unchanged. They apply an inhomogeneous process that reduces the diffusivity at those locations which have a larger likelihood to be edges. Let a gray-scale and 2-D (scalar-valued) image $u(x,y)$ be represented by a real-valued mapping in Ω C RxR . In Perona-Malik (PM) diffusion, the initial image $u_0(x,y)$ is modified through the anisotropic diffusion equation:

$$\partial u_t = div(g(\|\nabla u\|)\nabla u) \tag{1}$$

with $u(x,y,0) = u_0(x,y)$ as the initial condition. Here, *div* denotes the divergence operator, $u$ is the smoothened image at time step $t$, $\|\nabla u\|$ is the gradient magnitude of $u$, and $g(\|\nabla u\|)$ is the diffusivity function which is a nonnegative function of the magnitude of local image gradient $\|\nabla u\|$. The diffusivity function should be desirably a monotonically decreasing function approaching zero at infinity so that the diffusion process will take place only in the interior of regions and will not affect edges, where the gradient magnitude is significantly large. Another objective for the selection of $g(.)$ is to incur backward diffusion around large intensity transitions so that edges are sharpened, and to assure forward diffusion in smooth areas for noise removal. Two such diffusion coefficients suggested by Perona and Malik [4] are:

$$g(\|\nabla u\|) = \frac{1}{1 + \frac{\|\nabla u\|^2}{\delta^2}} \tag{2}$$

and

$$g(\|\nabla u\|) = \exp(-\frac{\|\nabla u\|^2}{\delta^2}) \tag{3}$$

where δ is the noise threshold which is a constant to be tuned for a particular application.

Diffusion can be modelled as an energy dissipating process [17]. Energy functional can be defined for smooth images as

$$E(u) = \int_{\Omega} f(\|\nabla u\|) d\Omega \tag{4}$$

where $f\|\nabla u\| \geq 0$ is an increasing function of $\|\nabla u\|$. This energy functional is a measure of image smoothness and its minimization is equivalent to smoothing. The minima of (4) are at some of its stationary points given by the Euler-Lagrange equation [18]

$$div\left[ f'(\|\nabla u\|) \frac{\nabla u}{\|\nabla u\|} \right] = 0 \tag{5}$$

Similar to gradient descent, (5) may be solved by the following parabolic equation

$$\partial u_t = div\left[ f'(\|\nabla u\|) \frac{\nabla u}{\|\nabla u\|} \right] \tag{6}$$

This is same as anisotropic diffusion (1) if

$$g(\|\nabla u\|) = \frac{f'(\|\nabla u\|)}{\|\nabla u\|} \tag{7}$$

In Perona-malik diffusion, the diffusion or the smoothing decreases as the gradient strength increases and the diffusion is stopped across edges. We next consider the decomposition of a diffusion process into diffusion along and across image edges to examine how anisotropic diffusion works locally to enhance/smear edges, or smooth/magnify noise.

**Orthogonal Decomposition of Anisotropic diffusion**

At a specific location $(x,y) \in \Omega$, the behaviour of anisotropic diffusion (6) is dependent on characteristics of $f[\|\nabla u(x,y)\|]$. For optimization purposes, $f[\|\nabla u(x,y)\|]$ is best characterized by the eigen structure of its Hessian matrix [17],

$$H(x,y) = \begin{bmatrix} \frac{\partial^2 f(\|\nabla u\|)}{\partial x^2} & \frac{\partial^2 f(\|\nabla u\|)}{\partial x \partial y} \\ \frac{\partial^2 f(\|\nabla u\|)}{\partial y \partial x} & \frac{\partial^2 f(\|\nabla u\|)}{\partial y^2} \end{bmatrix} \tag{8}$$

The eigen values of the *H* (*x*,*y*) are

$$\lambda_1(\|\nabla u\|)=\frac{f'(\|\nabla u\|)}{\|\nabla u\|}$$

and

$$\lambda_2(\|\nabla u\|)=f''(\|\nabla u\|) \tag{9}$$

From (7), $\lambda_1(\|\nabla u\|)=g(\|\nabla u\|)$ so anisotropic diffusion (6) can be expanded as,

$$\partial u_t=\lambda_1 D_o+\lambda_2 D_p \tag{10}$$

where

$$D_o=\frac{u_x{}^2 u_{yy}-2u_x u_y u_{xy}+u_y^2 u_{xx}}{u_x^2+u_y^2} \tag{11}$$

and

$$D_p=\frac{u_x^2 u_{xx}+2u_x u_y u_{xy}+u_y^2 u_{yy}}{u_x^2+u_y^2} \tag{12}$$

$D_o$ and $D_p$ are second order directional derivatives of *u* along directions orthogonal and parallel to the local gradient. Though the baseline fluctuations surrounding venous structures are suppressed by orthogonal decomposition, the filtered venous dips exhibit reduced amplitudes in comparison to the input. A more refined method uses the local structure coherence by controlling diffusion along directions of minimal and maximal curvatures.

**Directional Anisotropic Diffusion**

A more general approach for orthogonal decomposition of Perona Malik diffusion is to filter in the direction of gradient and the principal curvatures rather than filtering along directions parallel and perpendicular to the edges [15]. This is achieved by eigen value analysis of the Hessian matrix which is used to extract one or more principal directions of the local structure of the image. The Hessian matrix is computed for each pixel from the second order derivatives as,

$$H(x,y)=\begin{bmatrix} u_{xx} & u_{xy} \\ u_{yx} & u_{yy} \end{bmatrix} \tag{13}$$

The eigenvectors of the Hessian matrix computed for each pixel, point in the direction in which the second order image information takes extremal values and we call these directions as curvature directions. Using directional anisotropic diffusion, the individual slices are denoised by using a 3-D basis η, $e_1$ and $e_2$, which corresponds respectively to unit vectors along directions of the gradient, the maximal and minimal curvatures shown in Fig. 1.

**Fig. 1 here**

The denoised slices are obtained by

$$u = u_0 + \mu_1 u_\eta + \mu_2 u_{e1} + \mu_3 u_{e2} \tag{14}$$

where $u_0$ is the noisy slice, $u$ is the denoised slice, and $u_\eta$, $u_{e1}$ and $u_{e2}$ are the directional derivatives of image in the direction of gradient, maximum and minimum curvatures respectively. The $\mu$'s denote the respective diffusivity functions.

In the original work [15], a threshold $k$ is chosen based on the value of the gradient magnitude and supposes that $\|\nabla u\| > k$ in regions of frontier between different structures (or contours). If that condition is not met, the functions used are equivalent to the process of Perona and Malik. In the former case, diffusivity functions along maximum and minimum curvatures are modified to ensure that thediffusion along the direction of maximum curvature is stopped near the frontier ofthe vessels.

The choice of this threshold for detecting the upper limit of diffusion poses difficulties in the presence of spatially varying baseline. Even though, the venous dips are preserved by direct application of directional anisotropic filter, it does not provide background suppression which makes it difficult to trace venous structures located in grey matter, ventricle and tissues having relatively large iron content such as the caudate nucleus, thalamus and basal ganglia. Spatial adaptation of directional smoothing provides better venous contrast by increasing the amplitudes of venous dips along with background suppression. Detailed explanation of this technique follows.

## 2.2 Proposed Spatially adaptive diffusion

As discussed in section 1, intensity projected images can be modelled as the superposition of positive amplitude swings (MIP) or negative swings (mIP) on a baseline signal. The amplitude swings occur due to presence of vascular structures. Since the base line is more fluctuating for mIP images, we pre-process the component slices before projection. On the other hand, direct application of denoising on the projected image works better for maximum intensity projection.

For concomitant noise removal and venous enhancement, the mIP images are preprocessed by denoising individual slices along gradient and principle curvature directions. Spatial adaptation of the diffusivity functions further serves to increase contrast of vascular edges. A detailed explanation of the proposed algorithm follows.

### Extension to venous enhancement using SWI

Minimum Intensity Projection (mIP) consists of projecting pixels with minimum value along the volume of data. As discussed in section 1, mIP images have a fluctuating base line together with local dips. The is illustrated using a one dimensional cross-section of a mIP image in Fig. 2. Since the features of interest are the venous dips, it is required to retain, and preferably enhance the dip amplitudes along with simultaneous noise reduction in the surrounding base line regions.

**Fig. 2 here**

As discussed in section 2.1, this is achieved by spatially adapting the diffusivity function. After denoising each slice, a minimum intensity projection is first obtained. This is then point filtered using a mask derived from the phase image [19].

For denoising, we use a continuously varying diffusion function that depends on the structureness measure *C*, and second order directional derivatives. The structureness measure is used to distinguish background pixels and computed using [20],

$$C = \sqrt{u_{xx}^2 + u_{yy}^2} \tag{15}$$

Using the directional derivatives defined in section 2.1 and the structureness measure in Eq. (15), the diffusivity function is spatially adapted using

$$\mu = -\frac{1 - e^{-\alpha C u_e}}{1 + e^{-\alpha C u_e}} \tag{16}$$

which is computed individually for $u_\eta$, $u_{e1}$ and $u_{e2}$ respectively. Here, $\alpha$ is a constant which is strongly dependent on the SNR. *C* assumes low values in background regions and increases near the vessel edges, leading to increased slope of the diffusivity function in Eq. (16). Thus, the vessel edges can be automatically detected with the help of *C*, and obviates the need of a threshold. This serves as an advantage compared to the earlier method presented in [15].

The denoising process is repeated for each slice iteratively until it converges. The denoised slices are then minimum intensity projected. A block schematic of the proposed filter is shown in Fig. 3.

**Fig. 3 here**

**Extension to Magnitude flow enhancement in PC-PA MRA**

In multi-channel MRA, flow measurements are obtained using multiple measurements with changes in first order gradient moments [21]. Using either complex difference or phase difference approaches [21, 22], magnitude and directional flow information is obtained as maximum intensity projected images along X, Y and Z direction for each channel. Our filtering method is individually applied to the directionally combined images obtained by adding the X, Y and Z component flow images per channel. In addition to the proposed denoising scheme, we also provide a filter-synthesized scaling approach for improved phased array reconstruction of the magnitude flow images.

Consider a Phased Array (PA) comprising $k = 1, 2, . . . , N$ coils. The signal from coil *k* is separately digitized and stored as a set of complex numbers. The discrete Fourier transform of the data yields the single-coil complex image $Z_k$. The pixel-by-pixel modulus of $Z_k$ provides the single-coil magnitude image, $M_k$. The combined PA magnitude image is then given by [21]:

$$M = \sqrt{\sum_k \left( \frac{M_k}{\sigma_k} \right)^2} \tag{17}$$

where $\sigma_k$ is the spatially independent noise estimate for the *k*th coil-preamplifier-receiver chain, measured during pre-scan calibration. For majority of clinical scans, a pre-scan calibration may be difficult due to time constraints. In such situation, the proposed filter-synthesized scaling serves to provide high quality phased array reconstruction.

The local peaks of each magnitude image $M_k$ can be easily tracked using eigen vectors of minimum curvature. Hence, smoothing is performed only along the directions of the gradient and the minimum curvature. This is because for MIP, blood vessels occur as small elongated structures where the minimal curvature holds for the axis direction and the maximal curvature holds for the direction of the vessel cross section that is orthogonal to the gradient. Hence, denoising model for MIP takes the form

$$u = u_0 + \mu_1 u_\eta + \mu_3 u_{e2} \tag{18}$$

where the spatially adaptive diffusivity function is determined using

$$\mu = \frac{1 - e^{-\alpha C u_e}}{1 + e^{-\alpha C u_e}}, \text{ for } u_{e\min} < u_e < u_{e\max} \tag{19}$$

The directional derivatives at locations of vascular structures lie within a narrow range $[u_{e\min}, u_{e\max}]$. A sample histogram of $u_e$ is shown in Fig. 4. A threshold $u_{et}$ is chosen at both the knee points of the histogram so as to eliminate inclusion of noisy structural elements. When $u_{et}$ is chosen such that $\Pr(|u_e| > |u_{et}|) < 0.05$, the noise levels do not visually impair the image quality. In either direction, if $u_{et} > u_{e\max}$, the noise appears as black dots (pepper type) and for $u_{et} < u_{e\min}$, it takes form of white dots (salt type).

**Fig. 4 here**

Each filtered magnitude image $M_k^{'}$ is point filtered by a weight map prior to phased array reconstruction. The weights are determined using the ratio of sum of diffusion basis functions estimated during the final iteration of the filtering step applied to the projected image per channel, and the sum of diffusion basis functions obtained during filtering of phased array reconstructed $M_k$'s. This scaling is equivalent to Eq. (17) for spatially varying noise. The compensate phased array reconstruction is now performed using the spatially filtered $M_k^{''}$. The spatial filtering is mathematically represented using

$$M_k^{''} = M_k^{'} \times \frac{\sum_i \mu_{ik} u_{ik}}{\sum_i \mu_i u_i} \tag{20}$$

where $i$ denotes the direction of diffusion and $k$ is the channel number. The denominator term gives sum of diffusion basis functions obtained during filtering of unfiltered phased array reconstructed $M_k$'s. A detailed workflow is shown in Fig. 5.

**Fig. 5 here**

## 3. Results

The denoising is evaluated using flow compensated gradient echo images for SWI. The data sets used in this study have isotropic resolution of $(0.5mm \times 0.5mm \times 0.5mm)$ and acquired at two different echo times $T_E$=14.3ms and 17.3ms.

Fig. 6 shows the results of mIP applied to dataset #1 acquired at $T_E$=14.3ms. Panels (a)-(b) show the unfiltered mIP, (c) image filtered using spatially adaptive directional anisotropic diffusion with application of phase mask. The corresponding intensity profiles of a one-dimensional cross-section indicated by the red line in panel (a) are shown in (b1)-(c1). From the plots, it is clear that our method gives a less fluctuating base line with enhanced dip amplitudes for venous structures.

**Fig. 6 here**

Fig. 7 shows the results of mIP applied to dataset #2 acquired at $T_E$=17.3ms. Panels (a)-(b) show the unfiltered mIP, (c) image filtered using spatially adaptive directional anisotropic diffusion with application of phase mask. The corresponding intensity profiles of a one-dimensional cross-section indicated by the red line in panel (a) are shown in (b1)-(c1).

**Fig. 7 here**

Fig. 8 shows the results of mIP from anisotropic slices $(0.5mm \times 0.5mm \times 2mm)$ acquired at $T_E$=14.3ms. Panels (a)-(b) show the unfiltered mIP, (c) image filtered using spatially adaptive directional anisotropic diffusion with application of phase mask. The corresponding intensity profiles of a one-dimensional cross-section indicated by the red line in panel (a) are shown in (b1)-(c1).

**Fig. 8 here**

Fig. 9 shows the results of mIP from anisotropic slices acquired at $T_E$=17.3ms. Panels (a)-(b) show the unfiltered mIP, (c) image filtered using spatially adaptive directional anisotropic diffusion with application of phase mask. The corresponding intensity profiles of a one-dimensional cross-section indicated by the red line in panel (a) are shown in (b1)-(c1).

**Fig. 9 here**

Results of denoising applied to the MIP images for the four data sets discussed above are shown in Fig. 10. Panels (a1)-(a4) show the unfiltered MIP image, (b1-b4) the spatially adaptive directional anisotropic diffusion. Each row corresponds to isotropic slices acquired at $T_E$=14.3ms, 17.3ms and anisotropic slices acquired at $T_E$=14.3ms and 17.3ms respectively.

**Fig. 10 here**

Application of denoising applied to multi-channel PCMRA is illustrated in Fig. 11. MRA data is acquired using 6 head coils (channels) with $T_E$=9ms, $T_R$=56.70ms. Column-wise panels show unfiltered images, denoised images and images after compensation in each channel. The bottom row (a2)-(c2) show the combined images obtained after phased array combination. Right most column show that filter-synthesized scaling approach provides better background suppression.

**Fig. 11 here**

## 4. Discussions

We have proposed a method for denoising intensity projected images based on the concept of directional anisotropic diffusion. Since angiograms are obtained using intensity projections, the noise accumulated during acquisition process can cause swing amplitude reduction of vascular structures for both Maximum/Minimum Intensity Projected (M/mIP) images. For mIP images, vascular structures possess small swing amplitudes with reference to a base line. This poses difficulties to track vessels in presence of noise. A direct application of existing denoising methods are found to result in reduced amplitudes in the vascular regions. This can seriously affect the filtering performance since features of less prominent vessels are lost especially in mIP angiograms.

The spatially adaptive directional anisotropic diffusion uses a spatially varying diffusivity function $\mu$ for each direction along the gradient and the minimum and maximum curvature. A structureness measure $C$ is used to detect the vascular structures. The strength of the diffusion is modulated in such a way that the slope of the diffusivity function is increased in accordance with the structureness measure $C$. This slope factor is controlled by a measure $\alpha$ in the spatially adaptive diffusivity function. A decrease in the value of $\alpha$ leads to reduced contrast in the vascular regions. Increasing its value improves the continuity of vessels, but causes noise build-up. The variation of PSNR with $\alpha$ is shown in Fig. 12. The PSNR decreases with increasing value of $\alpha$, leading to less background suppression. So, we use hysteresis thresholding for improved results. For hysteresis thresholding, we choose two different $\alpha$'s with large enough difference in PSNR. The image corresponding to the larger PSNR is chosen as the reference image. For locations where $C$ exceeds a pre-determined threshold, the denoised intensities are estimated to be the ones obtained with higher value of $\alpha$. This technique yields in better continuity and contrast of vessels as shown in Fig. 13.

**Fig. 12, 13 here**

The proposed method is compared with performance of anisotropic diffusion proposed by Perona and Malik [3], directional anisotropic diffusion proposed by Krissian et.al [14] and the vesselness measure based method of Frangi et.al [15]. In particular, these comparisons are made using three quality measures: Peak Signal-to-Noise Ratio (PSNR) [23], Contrast ratio (CR) and Contrast per-pixel (C) [24].

PSNR is defined as

$$PSNR = 10\log_{10}\left(\frac{\max\left[u_{0ROI}\right]^2}{\frac{1}{MN}\sum_{m=1}^{M}\sum_{n=1}^{N}\left[u_{ROI}(m,n) - u_{0ROI}(m,n)\right]^2}\right) \tag{21}$$

where $M$, $N$ denotes the size of ROI, $u_{0ROI}$ and $u_{ROI}$ represent pixel values within the ROI of input and filtered images respectively. Contrast Ratio (CR) is defined as

$$CR_{ROI} = \frac{\max\left[u_{ROI}\right] - \min\left[u_{ROI}\right]}{\max\left[u_{ROI}\right] + \min\left[u_{ROI}\right]} \tag{22}$$

where $u_{ROI}$ denotes the filtered intensities within the chosen ROI. For an input image $u$ of size $M \times N$, the contrast per-pixel ($C$) is defined as the average difference in gray level between adjacent pixels and given by

$$C = \frac{\sum_{i=1}^{N}\sum_{j=1}^{M}\left(\sum_{(m,n)\in R(i,j)}\left|u(i,j) - u(m,n)\right|\right)}{MN} \tag{23}$$

where $R$ represents a local neighborhood of $(i,j)^{th}$ pixel. Increase in $C$ indicates average increase in contrast difference for each pixel compared to the neighbouring pixels. The results of comparison are summarized in Table 1 and 2. Major drawbacks of the anisotropic diffusion and directional anisotropic diffusion are the failure to detect less prominent vascular structures and background suppression. Even though the vesselness measure based method provides background suppression, it also introduces junction suppression as illustrated in Table 1. As evident from Table 1, the proposed algorithm is superior in terms of denoising and venous enhancement. In addition, it has the capability to detect the interconnected vascular structures. This is indicated by the red arrows in each figure. From Table 2, it is clear that the proposed method outperforms the existing methods in terms of PSNR, contrast ratio and contrast per pixel.

**Table. 1, 2 here**

## 5. Conclusion

In intensity projected images, noise is introduced during acquisition process. Due to this, vascular structures with low swing amplitudes become indistinguishable. Anisotropic diffusion is a well known method for denoising of intensity projected images. However, it does not have the potential to increase the swing amplitudes, especially for mIP image. So, a spatially adaptive directional anisotropic diffusion method is proposed for simultaneous denoising and venous enhancement. This method has a great potential to overcome the limitations of the existing denoising techniques applied to intensity projected images. A distinguishing feature is the introduction of a spatially varying diffusion function for controlling the diffusivity along the directions of gradient, minimum and maximum curvatures. The proposed method outperforms the existing methods in terms of venous enhancement, and junction suppression.

## Acknowledgements

The authors wish to thank the Department of Science and Technology (DST) of India for scholarship and operating funds.

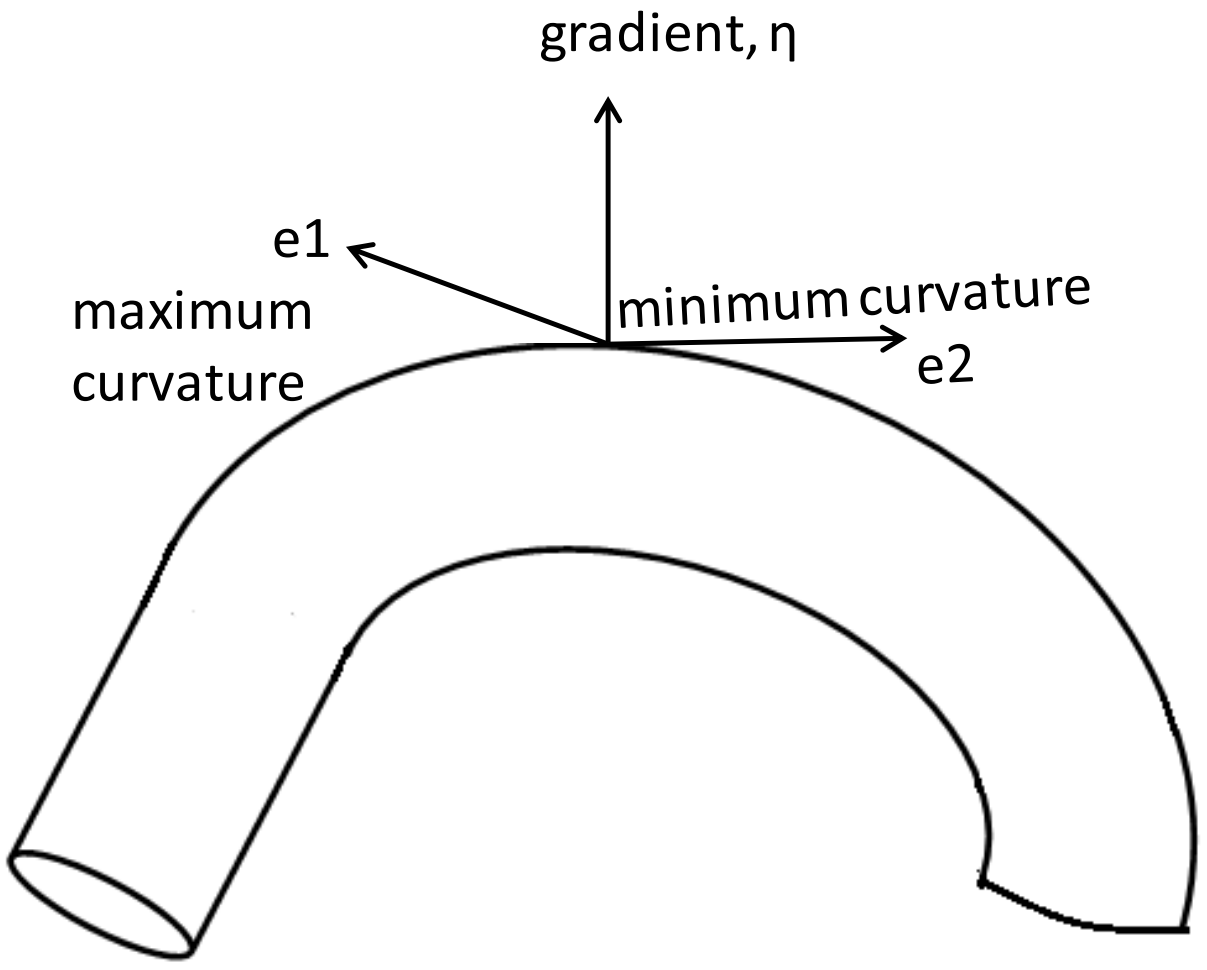

Fig. 1:- Illustration of the direction of the gradient, and the directions of the minimum and the maximum curvature.

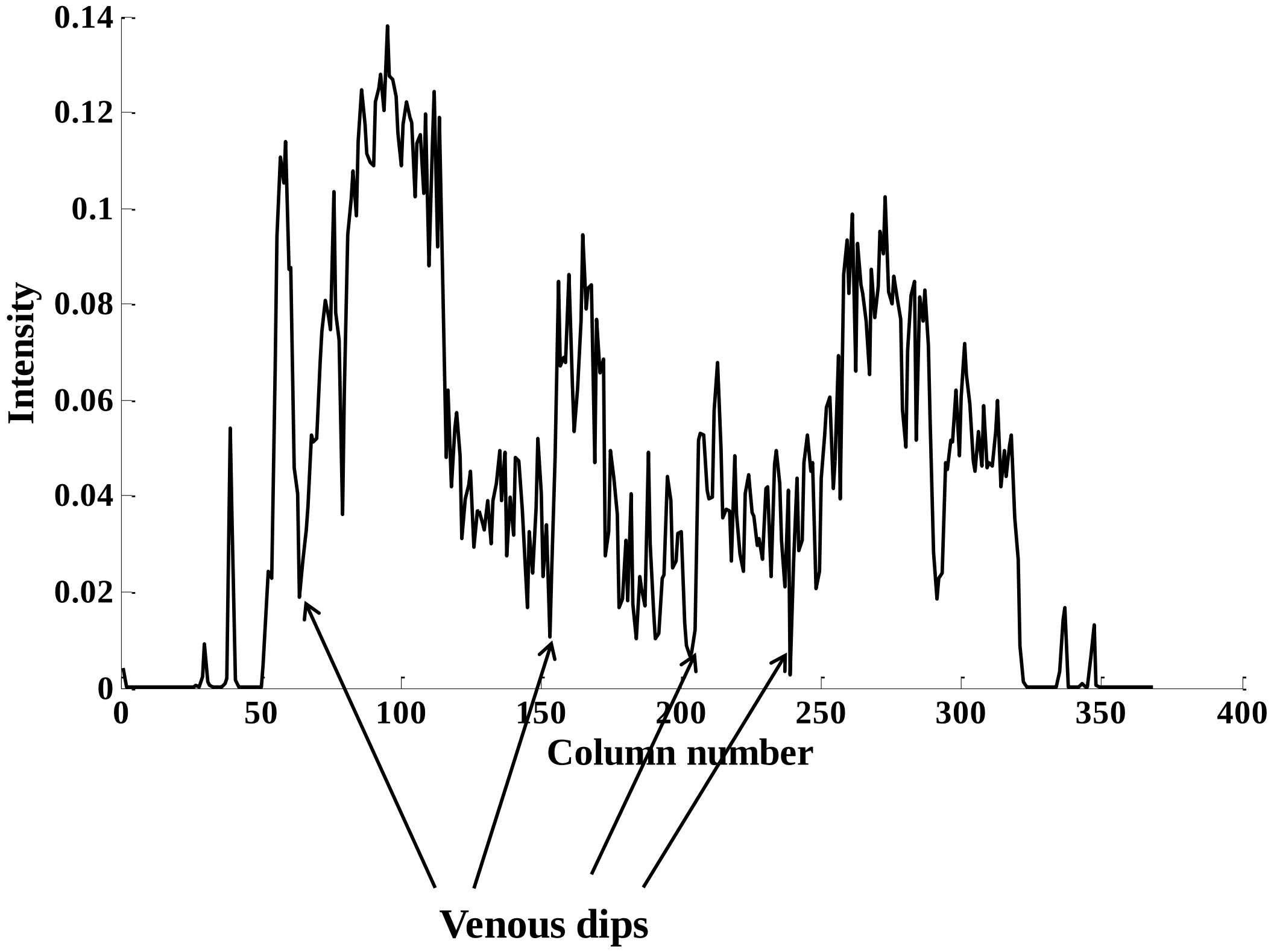

Fig. 2:- One dimensional cross-section of a mIP image

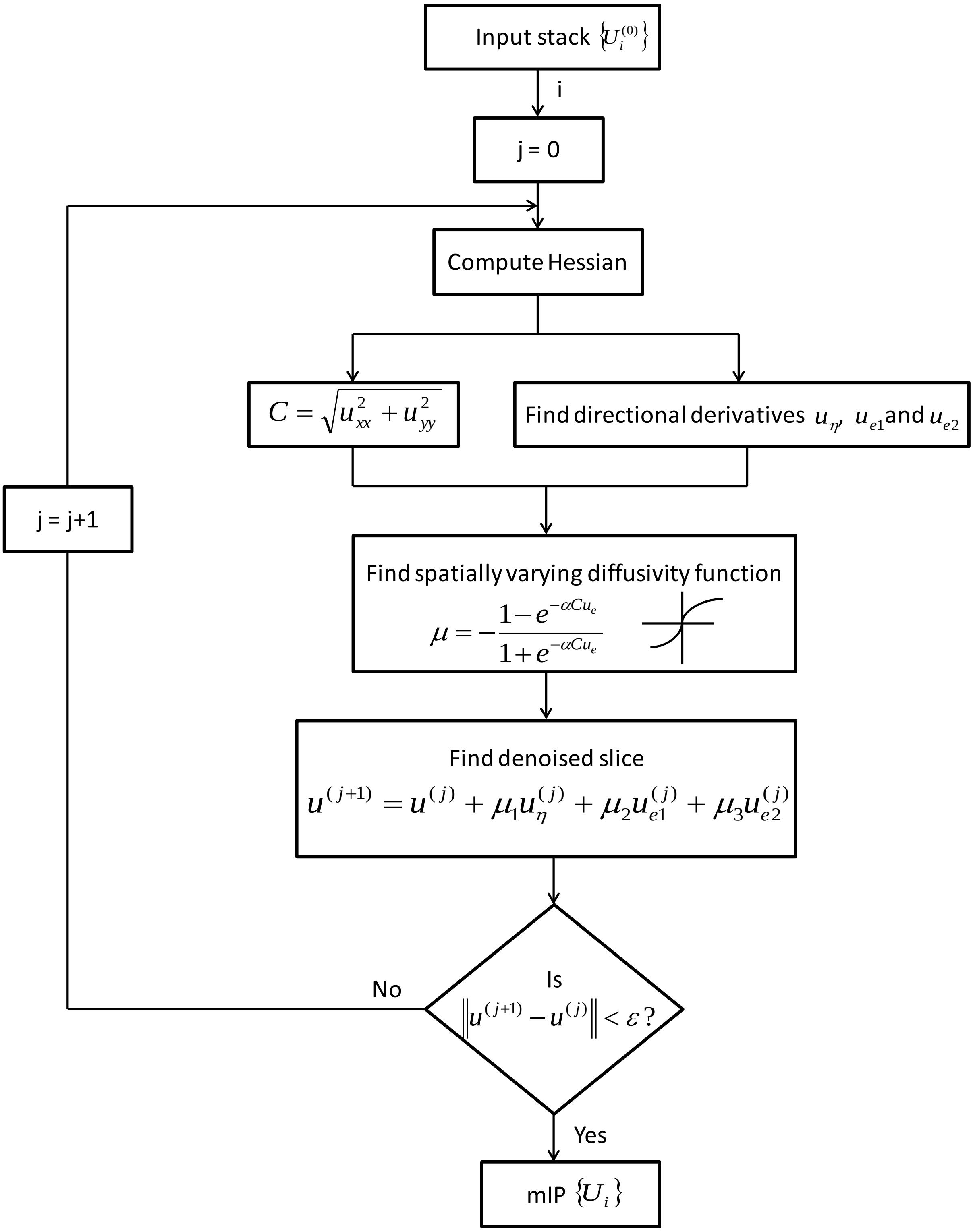

Fig. 3:- Workflow of the proposed method.

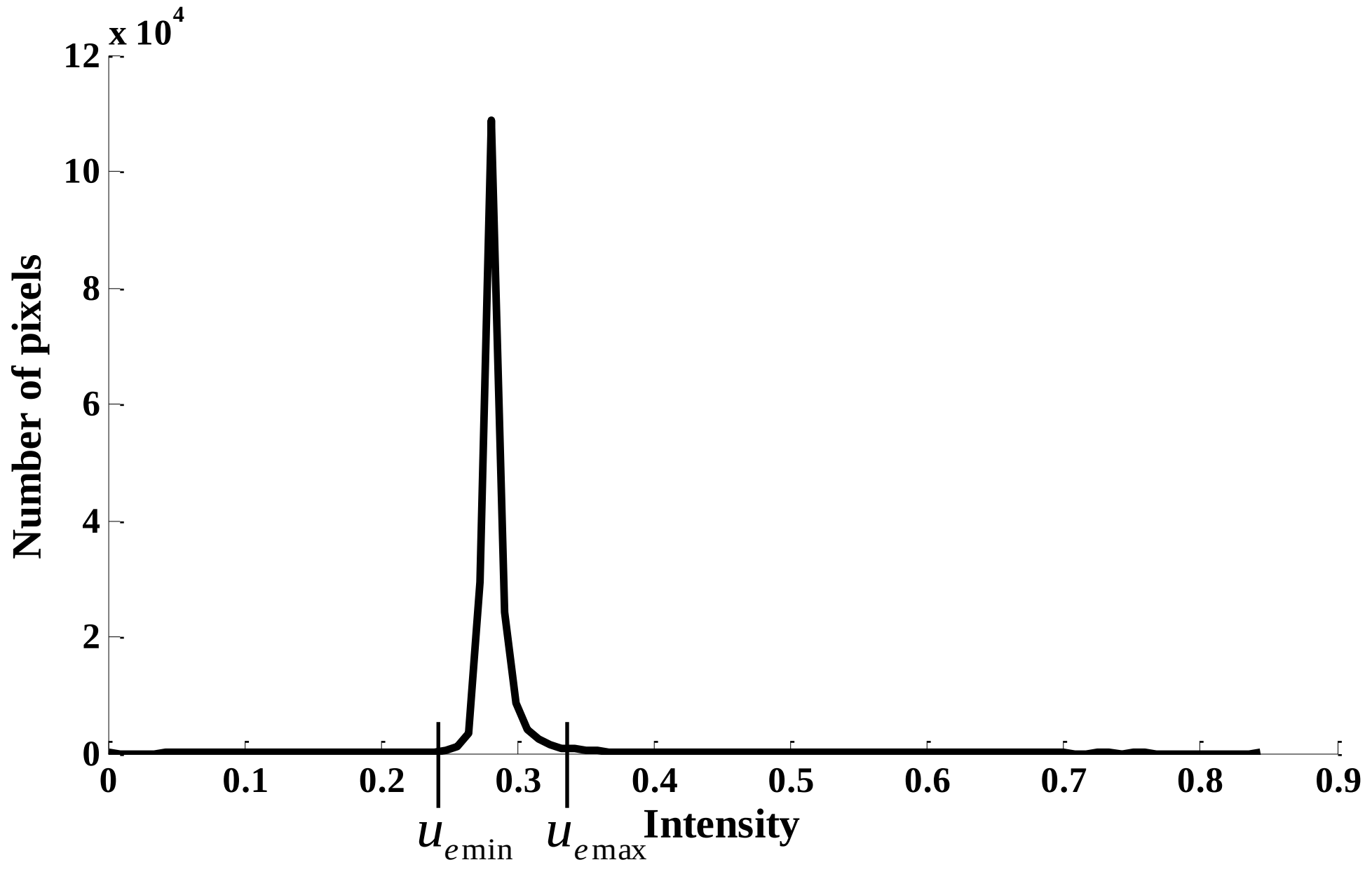

Fig. 4:- Intensity histogram of $u_{e3}$.

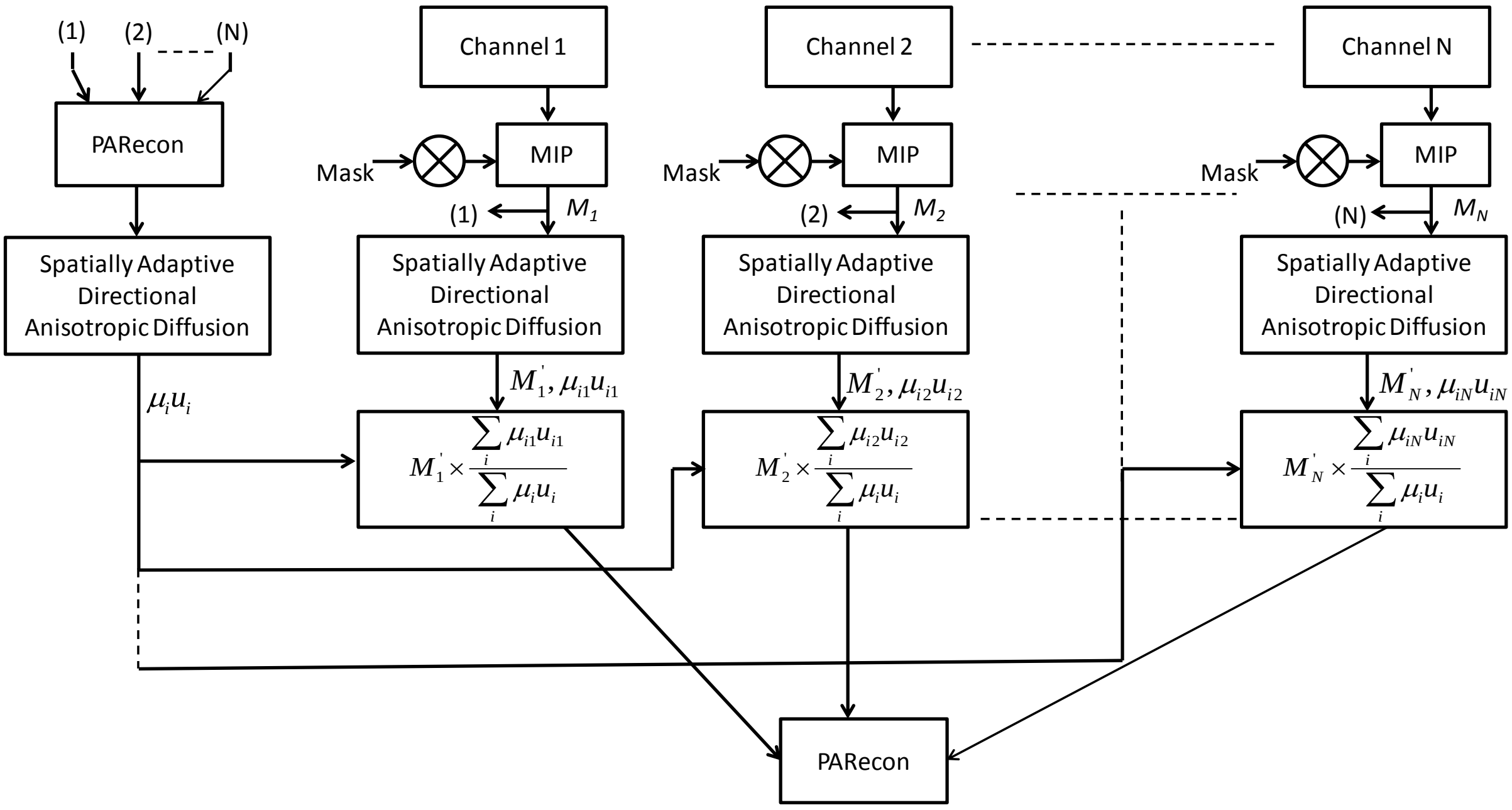

Fig. 5:- Filtering scheme for multichannel phase contrast angiogram using phased array reconstruction

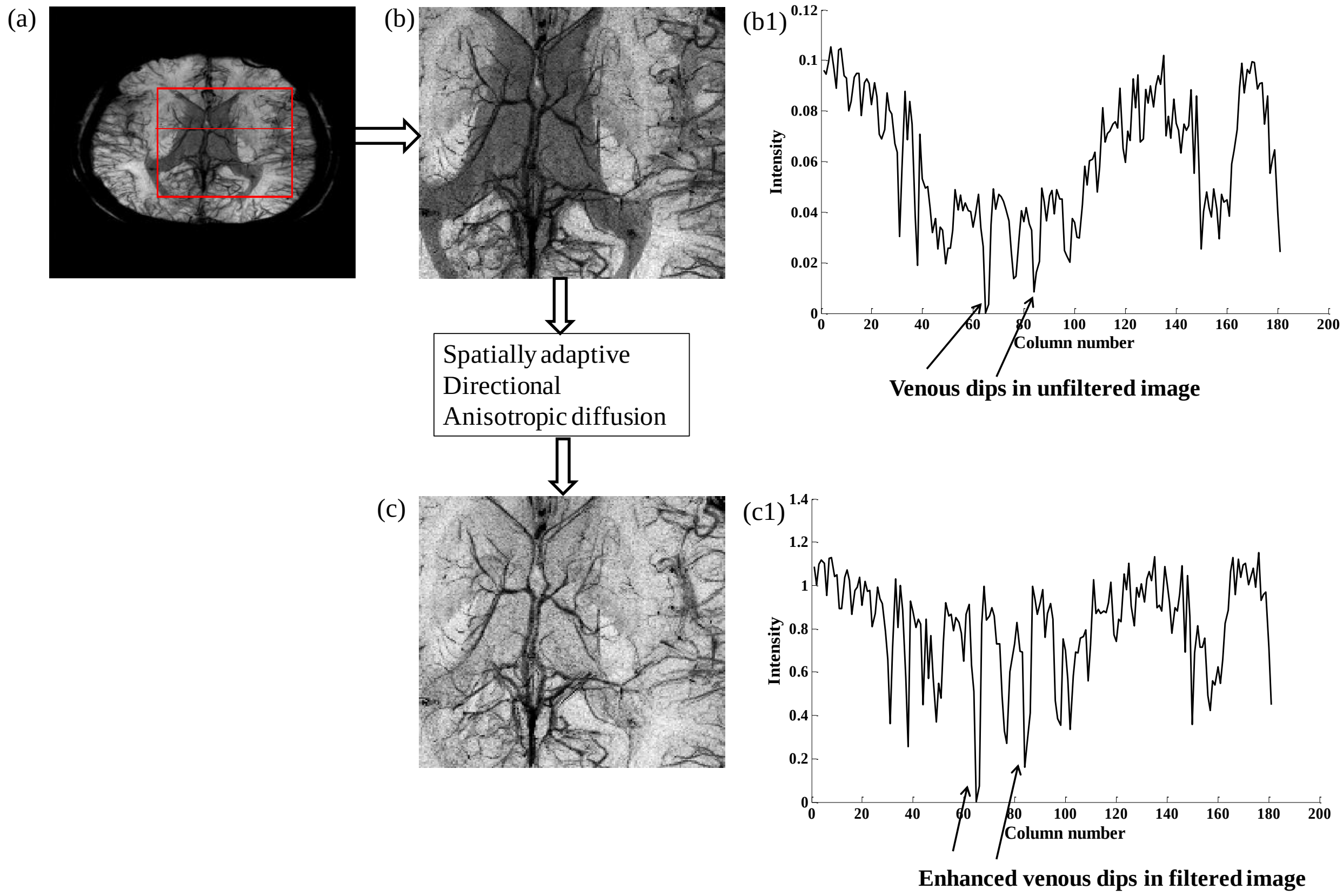

Fig. 6:- Isotropic slices acquired using $T_E$=14ms. (a)-(b) unfiltered mIP image with phase mask, (c) filtered image, (b1)-(c1) intensity profiles of (b)-(c) over a one dimensional cross-section indicated by red line in (a).

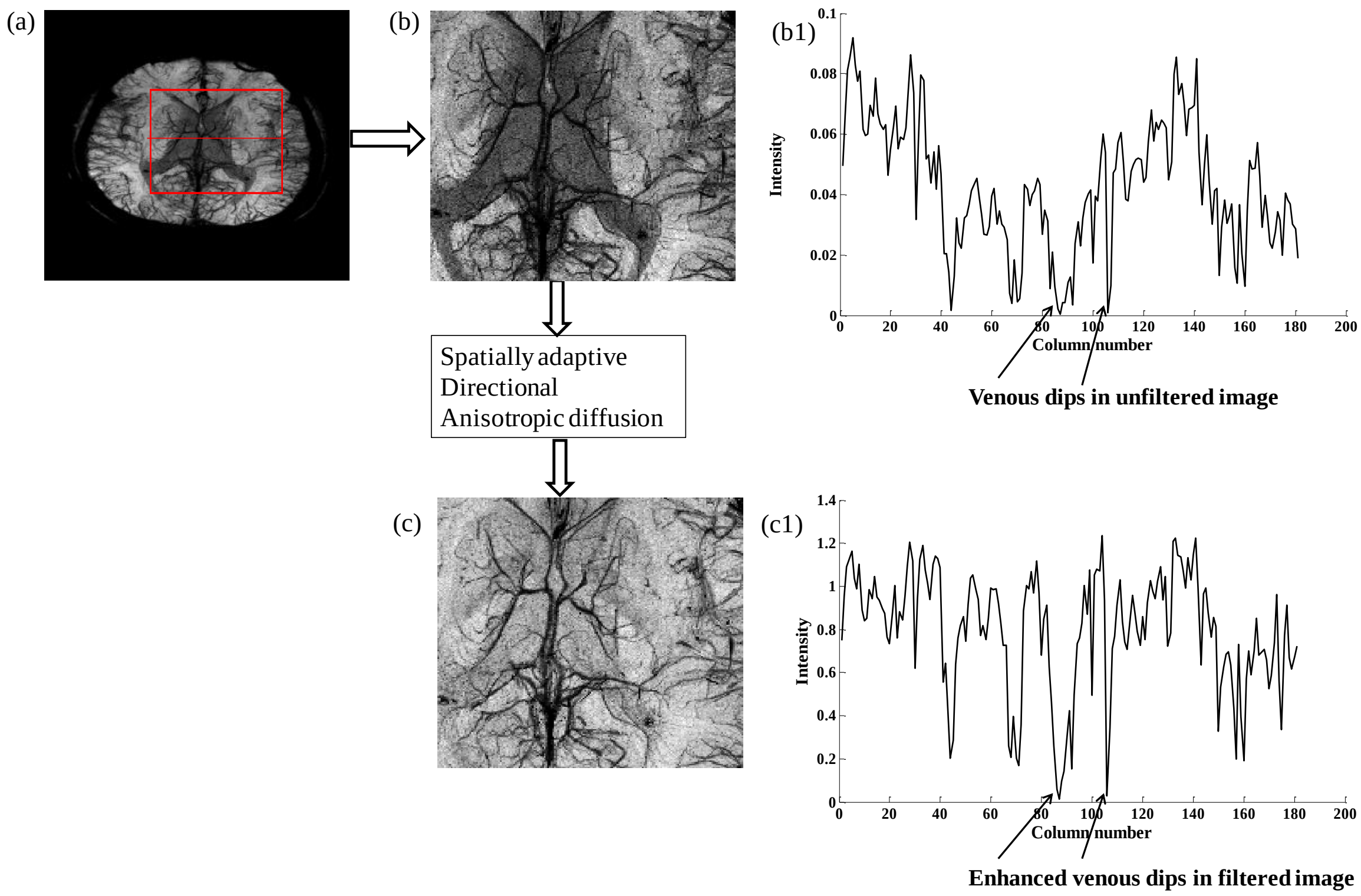

Fig. 7:- Isotropic slices acquired using $T_E$=17ms. (a)-(b) unfiltered mIP image with phase mask, (c) filtered image, (b1)-(c1) intensity profiles of (b)-(c) over a one dimensional cross-section indicated by red line in (a).

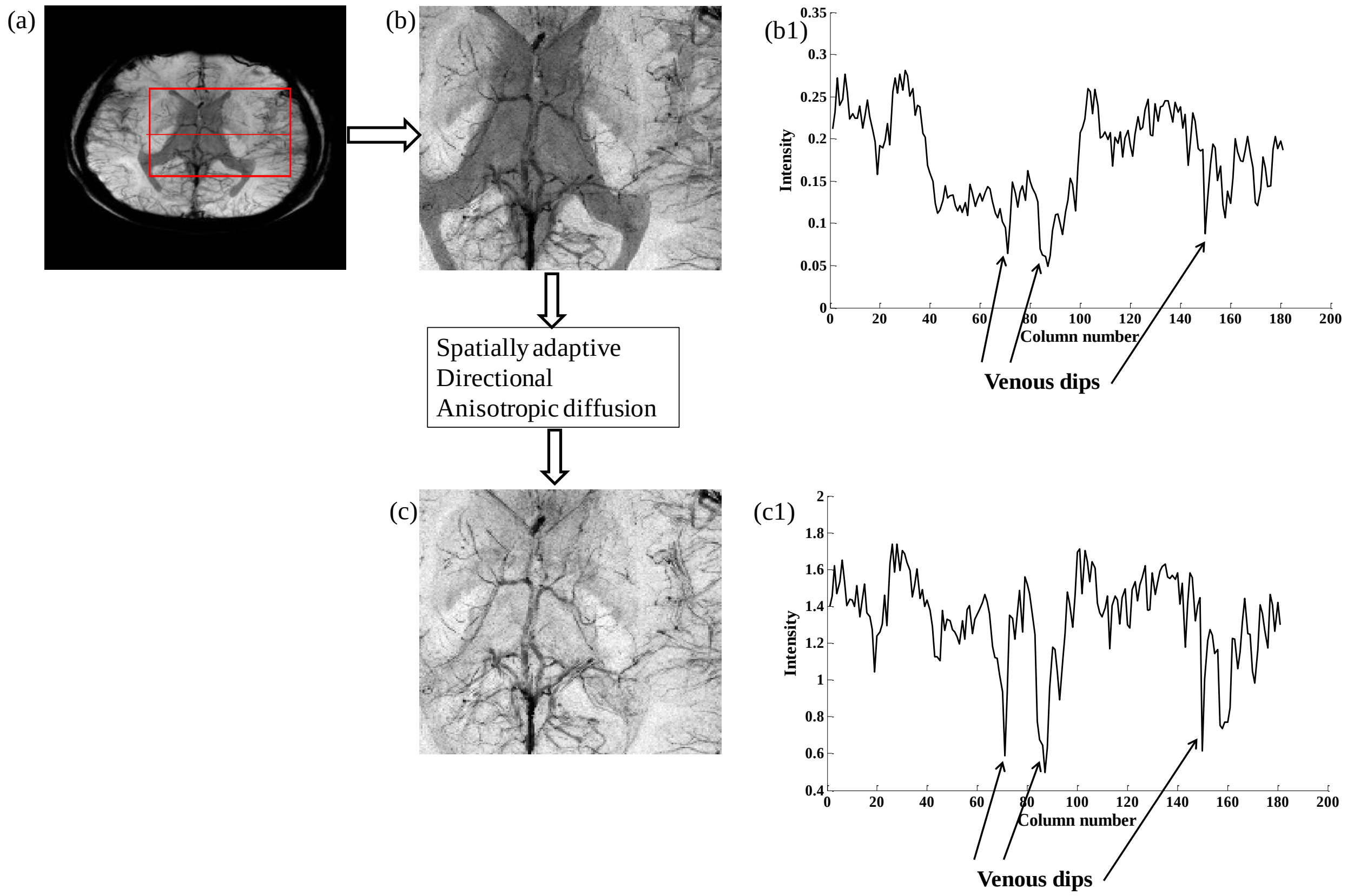

Fig. 8:- Anisotropic slices acquired using $T_E$=14ms. (a)-(b) unfiltered mIP image with phase mask, (c) filtered image, (b1)-(c1) intensity profiles of (b)-(c) over a one dimensional cross-section indicated by red line in (a).

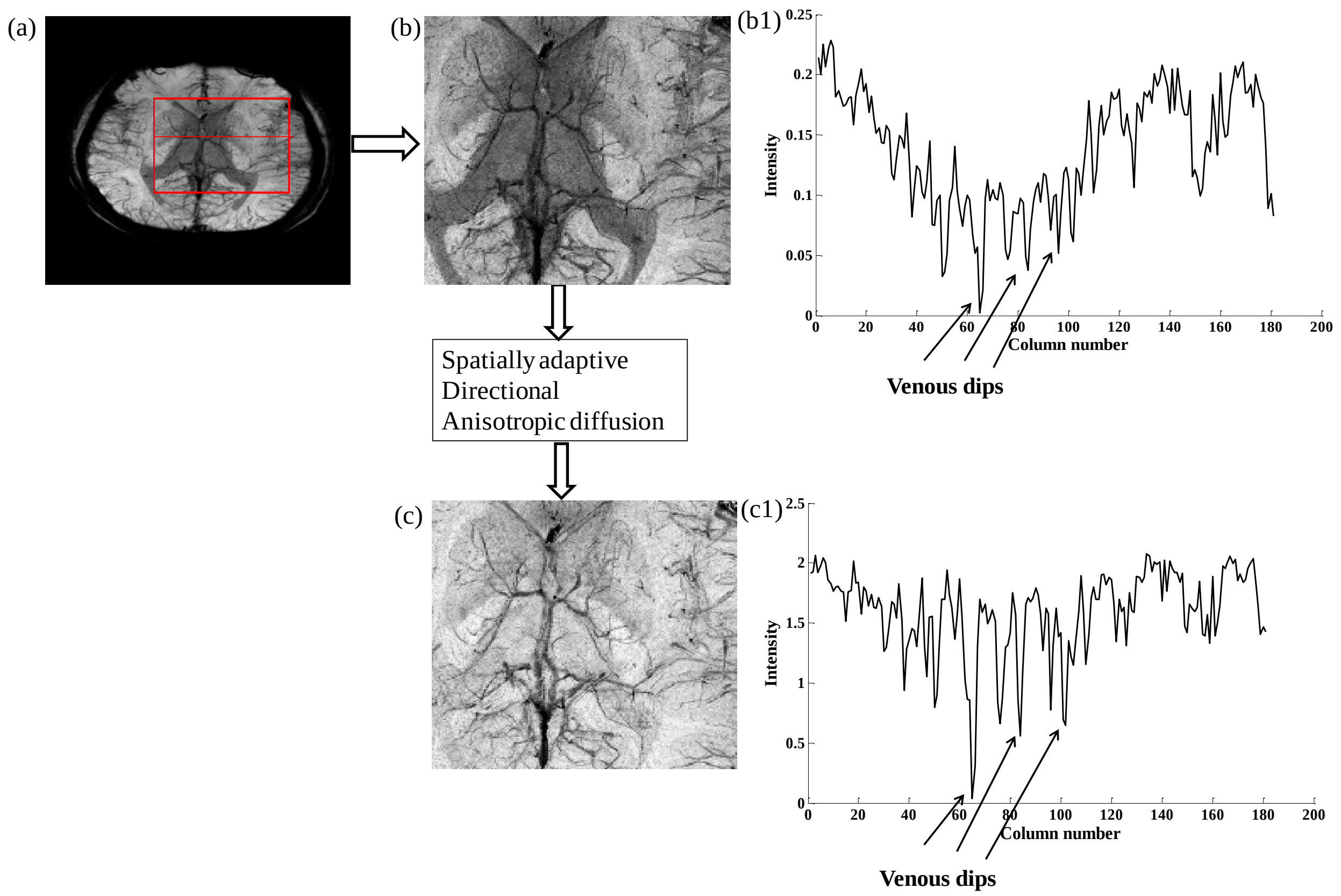

Fig. 9:- Anisotropic slices acquired using $T_E$=17ms. (a)-(b) unfiltered mIP image with phase mask, (c) filtered image, (b1)-(c1) intensity profiles of (b)-(c) over a one dimensional cross-section indicated by red line in (a).

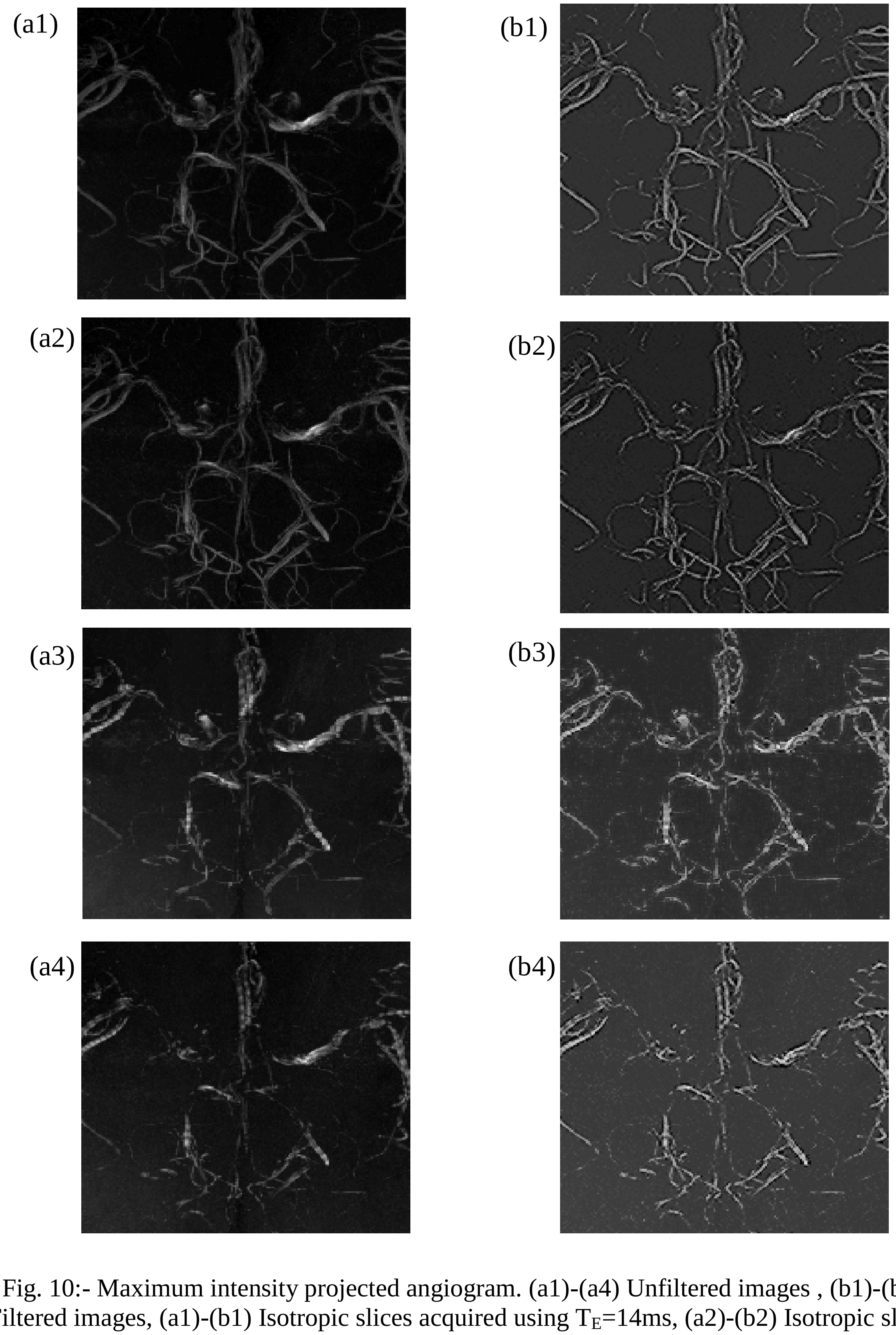

Fig. 10:- Maximum intensity projected angiogram. (a1)-(a4) Unfiltered images , (b1)-(b4) Filtered images, (a1)-(b1) Isotropic slices acquired using $T_E$=14ms, (a2)-(b2) Isotropic slices acquired using $T_E$=17ms, (a3)-(b3) Anisotropic slices acquired using $T_E$=14ms (a4)-(b4) Anisotropic slices acquired using $T_E$=17ms.

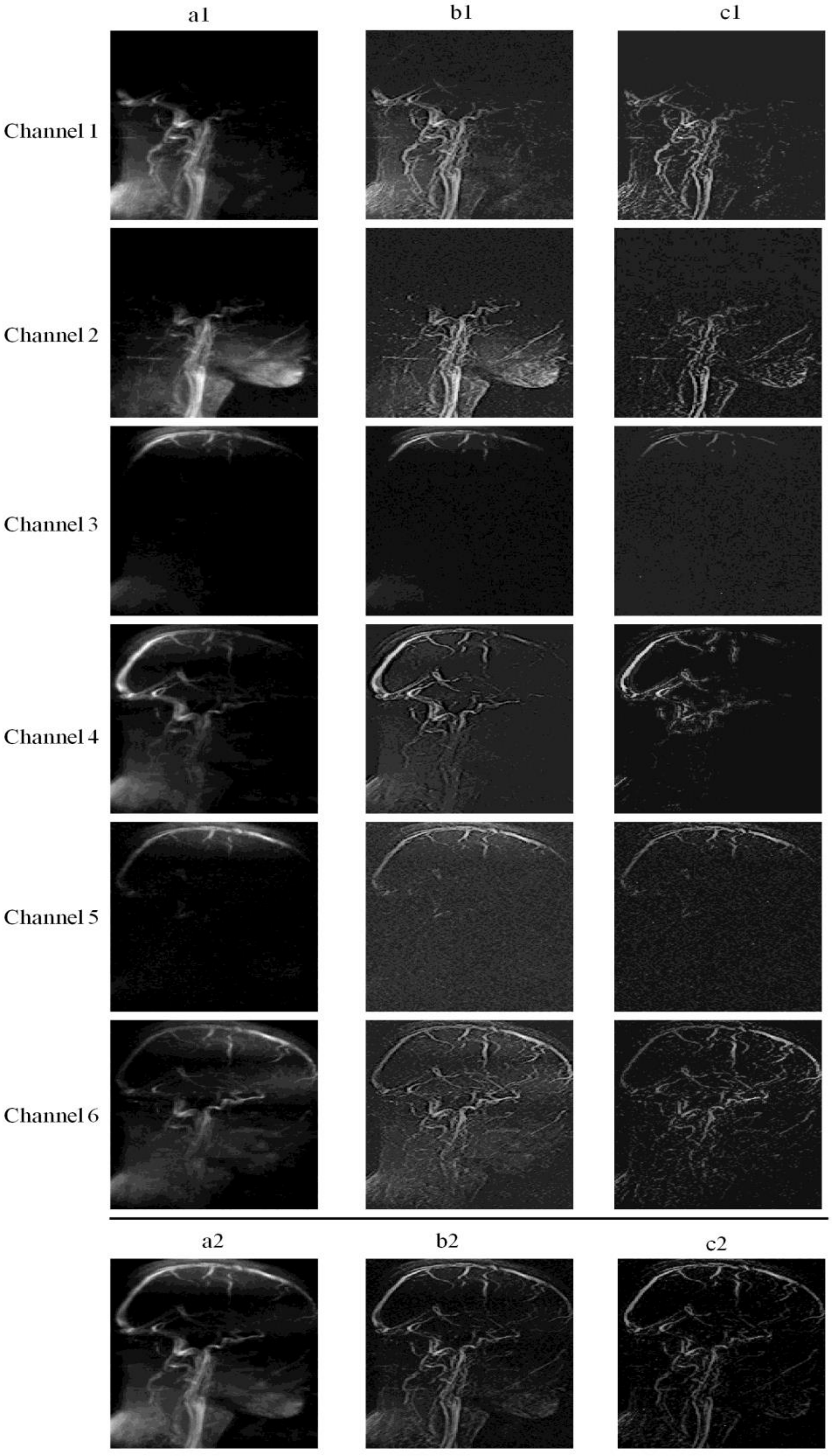

Fig. 11:- Multi-channel Phase Contrast MR angiogram. (a1)-(c1) Individual channel images, (a2)-(c2) Phased Array reconstructed images, (a1)-(a2) Unfiltered image, (b1)-(b2) Filtered image, (c1)-(c2) Image reconstructed using improved phase array reconstruction using filter-synthesized scaling approach.

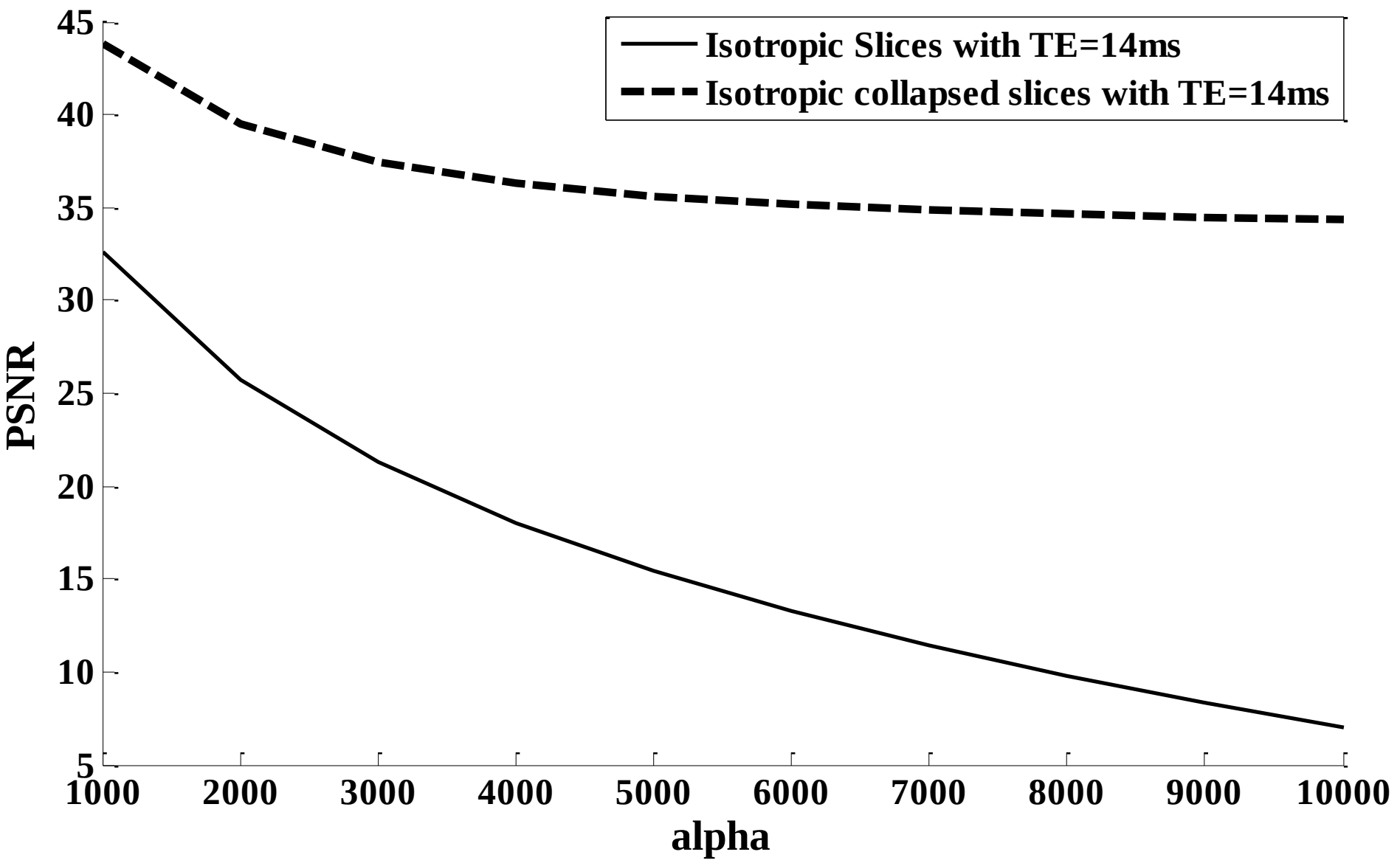

Fig. 12:- Variation of PSNR with $\alpha$

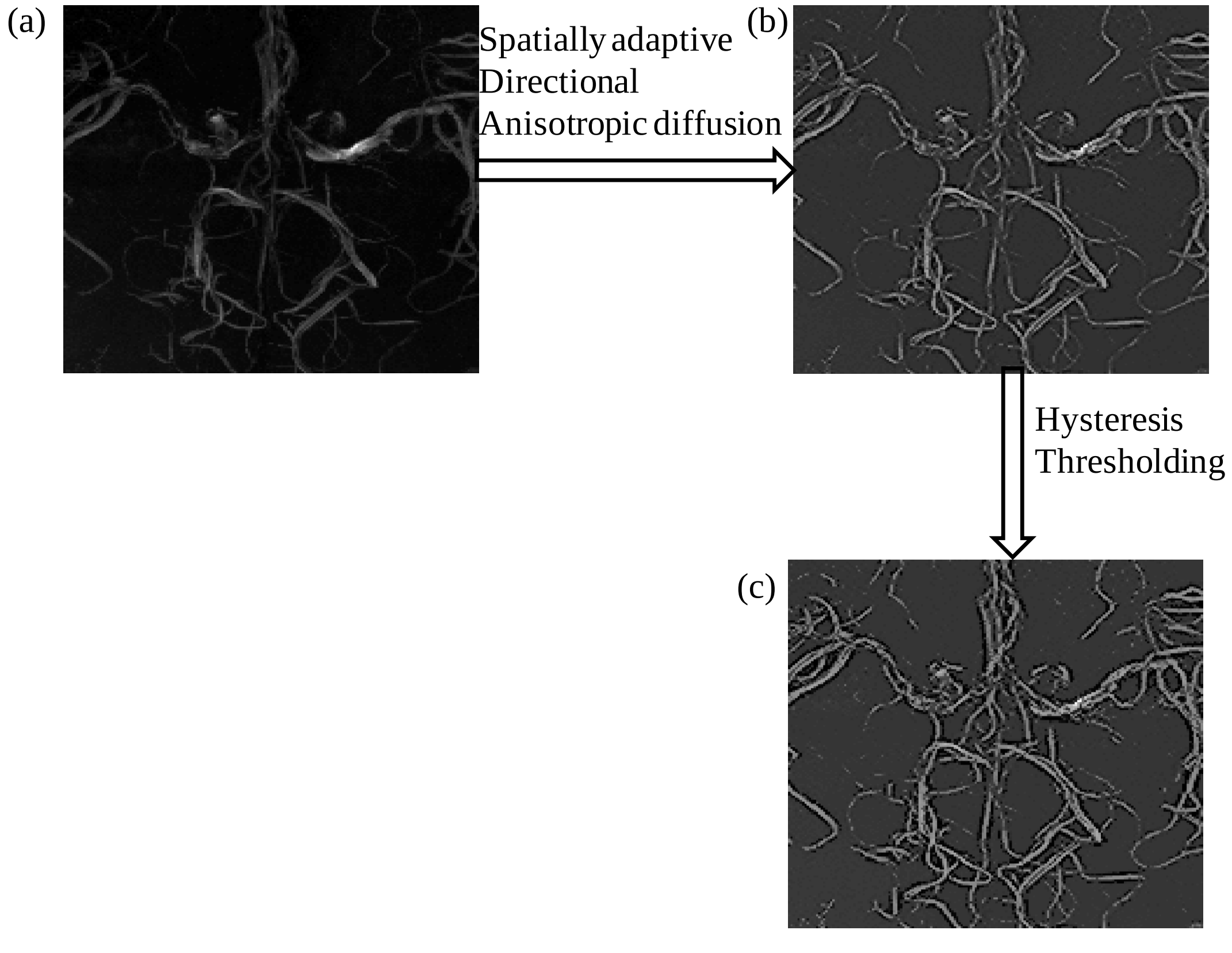

Fig. 13:- Isotropic slices acquired using $T_E$=14ms. (a)unfiltered MIP image, (b)filtered image, (c)result of hysteresis thresholding

Table 1: Comparison of the proposed method with existing techniques

| | | |
|---|---|---|
| mIP image | 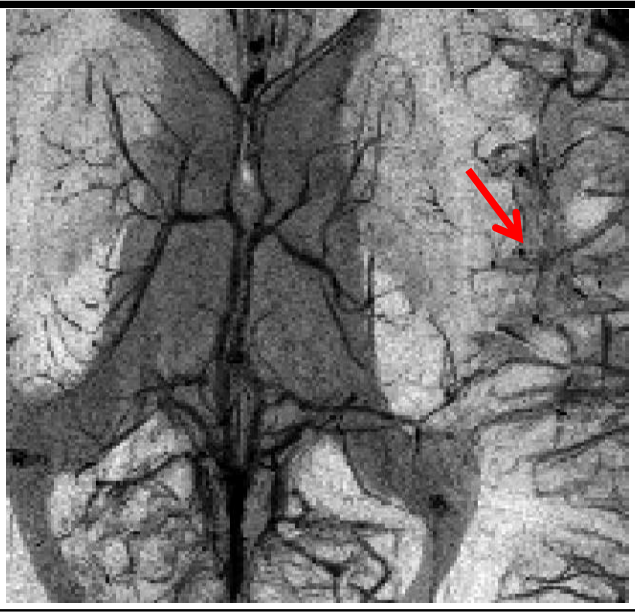 | 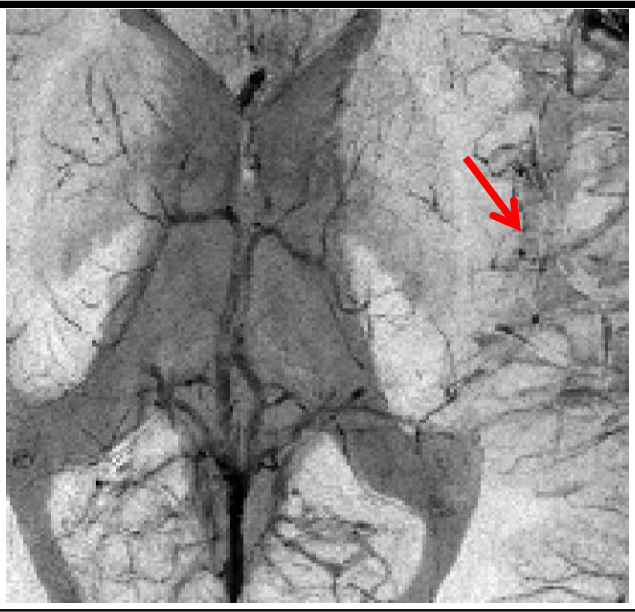 |
| Anisotropic diffusion filtered | 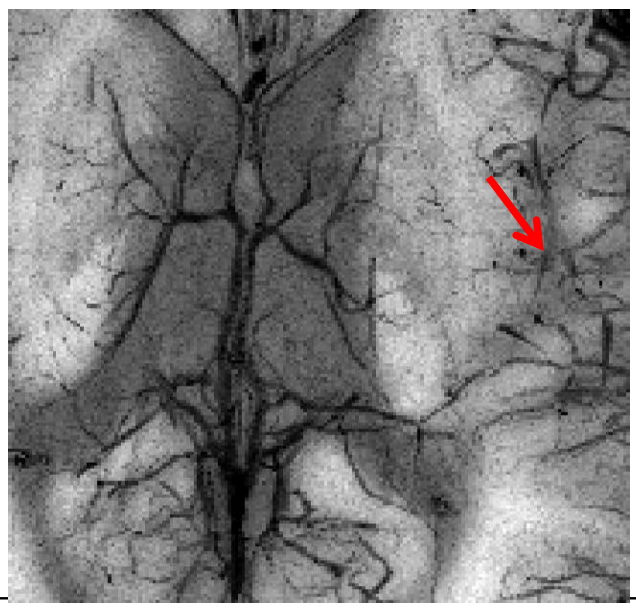 | 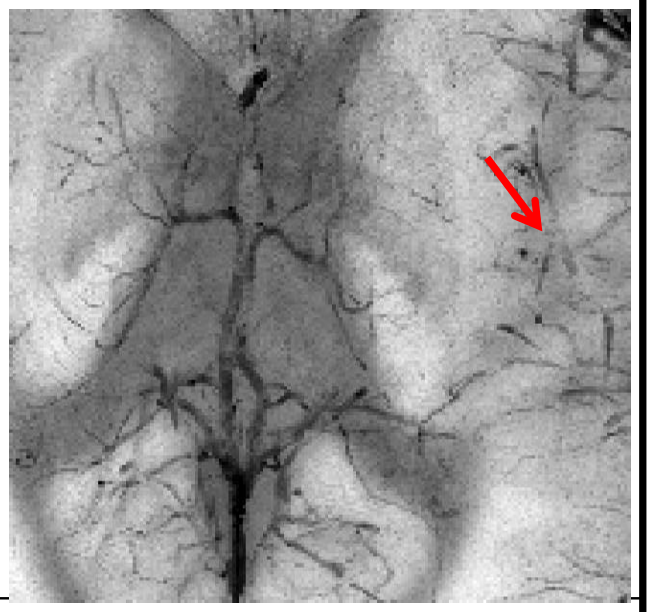 |
| Directional anisotropic diffusion filtered | 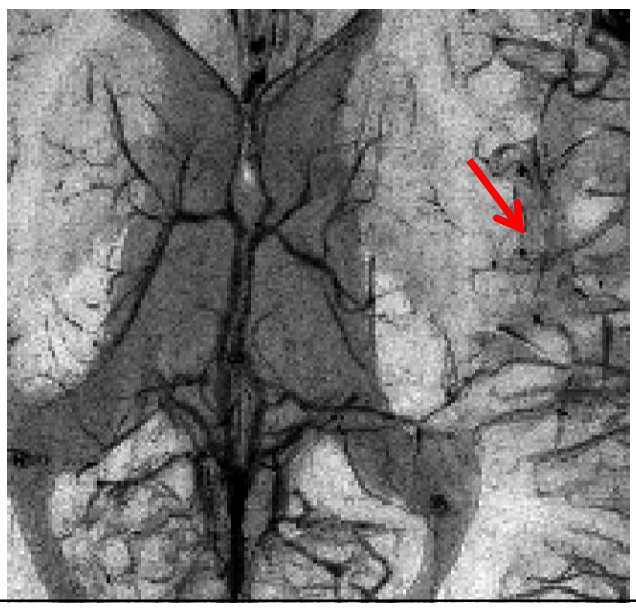 | 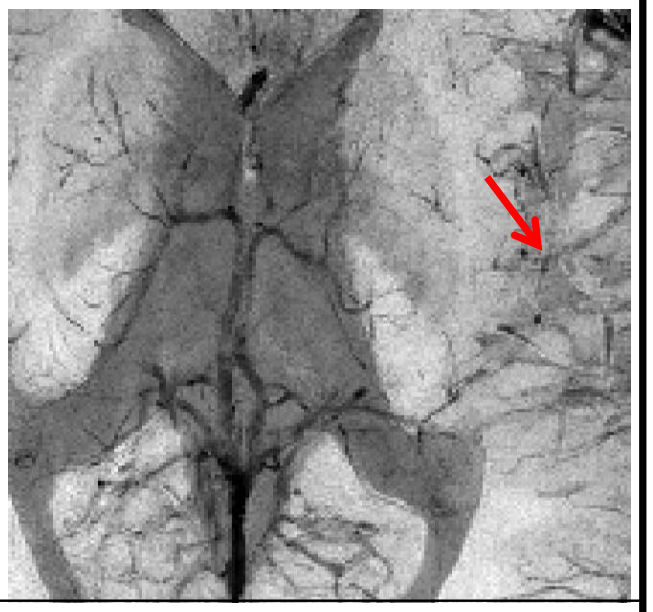 |
| Multiscale vessel enhancement based filtered | 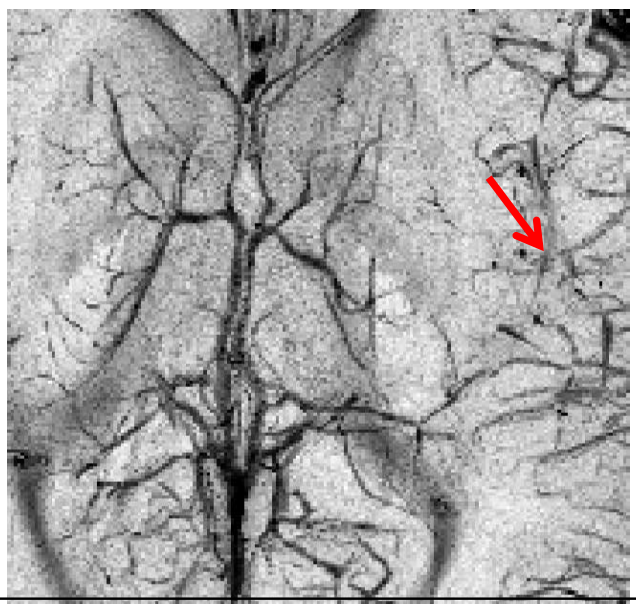 | 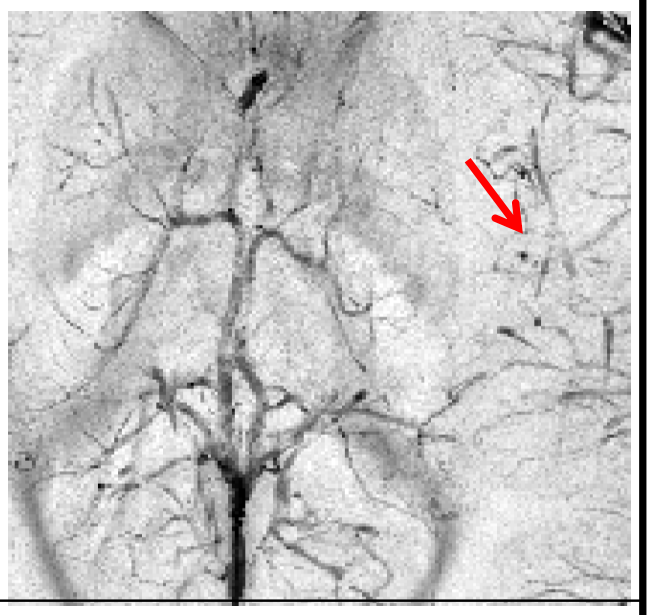 |
| Spatially adaptive directional anisotropic diffusion filtered | 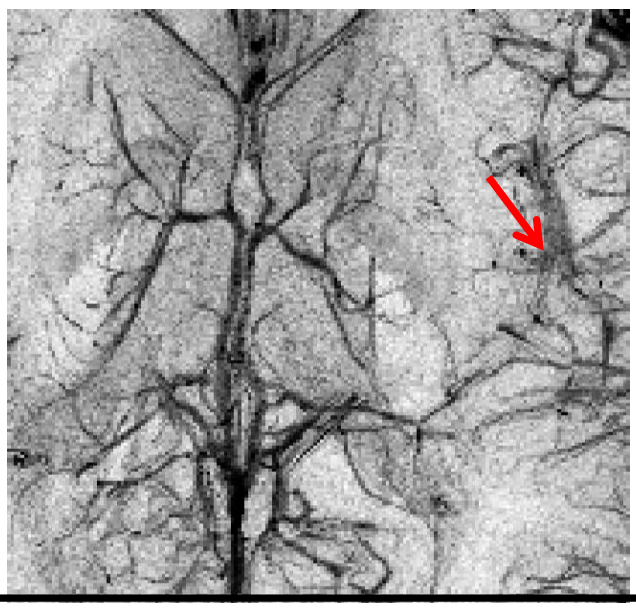 | 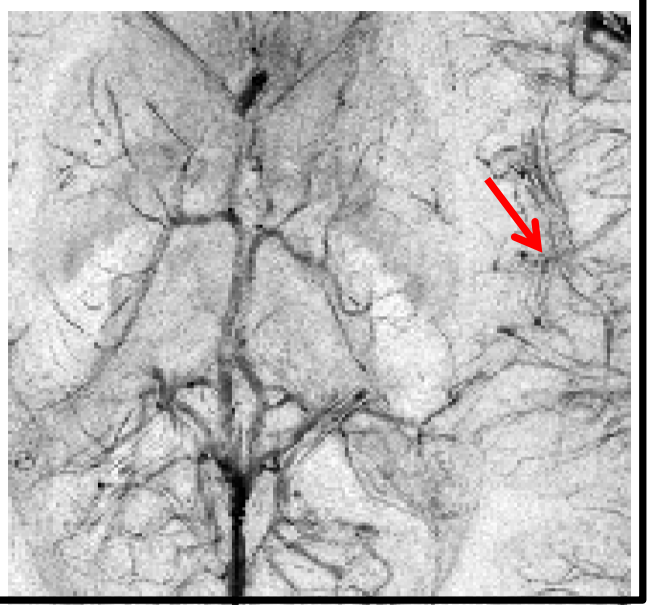 |

Table 2: Performance comparison of the proposed method with state-of-the-art methods

| Method | PSNR | Contrast Ratio (CR) | Contrast per pixel (C) |
|---|---|---|---|
| Anisotropic diffusion filter | 21.2613 | 0.8061 | 0.7375 |
| Directional anisotropic diffusion filter | 23.1477 | 0.9007 | 0.8974 |
| Multiscale vessel enhancement filter | 21.0120 | 0.8059 | 1.3394 |
| Spatially adaptive directional anisotropic diffusion | 24.7173 | 0.9177 | 1.3791 |